%% file: ijcai24.tex
\documentclass{article}
\usepackage{ijcai24}

\usepackage{times}
\usepackage{soul}
\usepackage{url}
\usepackage[hidelinks]{hyperref}
\usepackage[utf8]{inputenc}
\usepackage[small]{caption}
\usepackage{graphicx}
\usepackage{amsmath}
\usepackage{amsthm}
\usepackage{booktabs}
\usepackage{algorithm}
\usepackage{algorithmic}
\usepackage[switch]{lineno}
\usepackage{multirow}
\usepackage[table]{xcolor}

\title{RWKV-TS: Beyond Traditional Recurrent Neural Network for Time Series Tasks}

\author{
Haowen Hou$^1$
\and
F. Richard Yu$^1$
\affiliations
$^1$Guangming Laboratory \\
\emails
houhaowen@gml.ac.cn,
yufei@gml.ac.cn 
}

\begin{document}

\maketitle

\begin{abstract}
Traditional Recurrent Neural Network (RNN) architectures, such as LSTM and GRU, have historically held prominence in time series tasks. However, they have recently seen a decline in their dominant position across various time series tasks. As a result, recent advancements in time series forecasting have seen a notable shift away from RNNs towards alternative architectures such as Transformers, MLPs, and CNNs.
To go beyond the limitations of traditional RNNs, we design an efficient RNN-based model for time series tasks, named RWKV-TS, with three distinctive features:
(i) A novel RNN architecture characterized by $O(L)$ time complexity and memory usage.
(ii) An enhanced ability to capture long-term sequence information compared to traditional RNNs.
(iii) High computational efficiency coupled with the capacity to scale up effectively.
Through extensive experimentation, our proposed RWKV-TS model demonstrates competitive performance when compared to state-of-the-art Transformer-based or CNN-based models. Notably, RWKV-TS exhibits not only comparable performance but also demonstrates reduced latency and memory utilization.
The success of RWKV-TS encourages further exploration and innovation in leveraging RNN-based approaches within the domain of Time Series. The combination of competitive performance, low latency, and efficient memory usage positions RWKV-TS as a promising avenue for future research in time series tasks. Code is available at:\href{https://github.com/howard-hou/RWKV-TS}{ https://github.com/howard-hou/RWKV-TS}
\end{abstract}

\section{Introduction}

\input{sections/Introduction}

\section{Related Work}

\input{sections/Related_Work}

\section{Model Architecture}

\input{sections/Model_Architecture}

\section{Experiment}

\input{sections/Experiment}

\section{Conclusion}

\input{sections/Conclusion}

\section*{Ethical Statement}

There are no ethical issues.

\section*{Acknowledgments}
Thanks to Peng Bo, the author of RWKV, for participating in the discussion and providing valuable suggestions for modifications. 

\clearpage
\bibliographystyle{named}
\bibliography{ijcai24}

\clearpage
%%%%%%%%%%%%%%%%%%%%%%%%%%%%%%%%%%%%%%%%%%%%%%%%%%%%%%%%%%%%%%%%%%%%%%%%%%%%%%%
%%%%%%%%%%%%%%%%%%%%%%%%%%%%%%%%%%%%%%%%%%%%%%%%%%%%%%%%%%%%%%%%%%%%%%%%%%%%%%%
% APPENDIX
%%%%%%%%%%%%%%%%%%%%%%%%%%%%%%%%%%%%%%%%%%%%%%%%%%%%%%%%%%%%%%%%%%%%%%%%%%%%%%%
%%%%%%%%%%%%%%%%%%%%%%%%%%%%%%%%%%%%%%%%%%%%%%%%%%%%%%%%%%%%%%%%%%%%%%%%%%%%%%%
\input{sections/Appendix}

\end{document}

%% file: sections/Introduction.tex
\begin{figure}[htbp]
    \centering
    \includegraphics[width=0.45\textwidth]{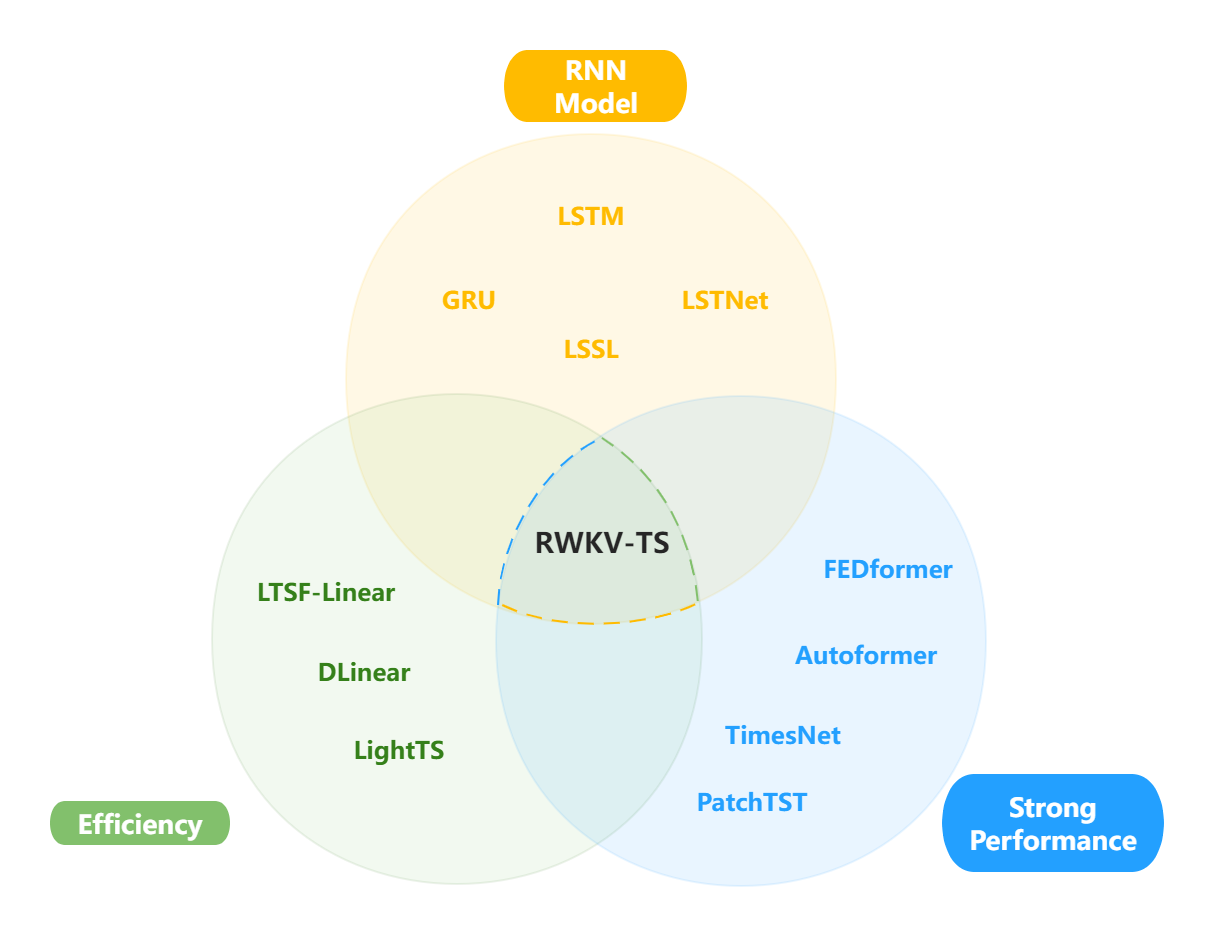}
    \caption{RWKV-TS is a time-series RNN-based model that achieves both strong performance and efficiency simultaneously. In contrast, other RNN models are considered to perform poorly in both aspects for time-series tasks.}
    \label{fig:RWKV-TS advantage}
\end{figure}

Time series analysis is a fundamental domain in machine learning with diverse applications ranging from financial forecasting to environmental prediction. Historically, traditional Recurrent Neural Networks (RNNs) have been the cornerstone for modeling sequential data due to their inherent ability to capture temporal dependencies. 
However recently, traditional RNNs have gradually lost their dominant position in time series tasks, and the leaderboard\footnote{https://github.com/thuml/Time-Series-Library} for time series has also been taken over by alternative architectures such as Transformers, Multi-Layer Perceptrons (MLPs), and Convolutional Neural Networks (CNNs). 

Traditional Recurrent Neural Networks (RNNs), whether in their vanilla form or as variants like Long Short-Term Memory (LSTM)\cite{lstm} and Gated Recurrent Unit (GRU) \cite{gru}, exhibit the following limitations: 
\begin{itemize}
    \item The vanishing/exploding gradient problem constrains the ability to capture information in long sequences. As the sequence length exceeds 100, the capacity to capture information rapidly diminishes\cite{Kaplan2020ScalingLF}. 
    \item The inability to perform parallel computations results in low computational efficiency and hinders the ability to scale up. 
    \item The step-by-step prediction approach leads to error accumulation and subpar inference speeds\cite{zhou2021informer}. 
\end{itemize}
Recent literature\cite{zhou2021informer,zhou2022fedformer} has argued that RNNs may no longer be the optimal choice for time series tasks involving the modeling of long-term dependencies. This leads us to ponder: 
\textit{are RNNs truly unsuitable for handling time series tasks with long-term dependencies? }

\textit{The answer may well be No.} To address the limitations of traditional RNNs, we designed an efficient RNN-based model for time series tasks, which we have named RWKV-TS, distinguished by three key features: 
\begin{enumerate}
    \item A novel RNN architecture that is characterized by $O(L)$ time complexity and memory usage.
    \item  An improved capacity to capture long-term sequence information, surpassing that of traditional RNNs. 
    \item High computational efficiency, coupled with the ability to scale up effectively.
\end{enumerate}

RWKV-TS has demonstrated competitive performance comparable to state-of-the-art (SOTA) models across various time-series prediction tasks such as Long-term Forecasting, Short-term Forecasting, Imputation, Anomaly Detection, Classification, Few-shot, effectively validating the potential of linear RNN models in time-series tasks. As shown in Figure \ref{fig:RWKV-TS advantage}, RWKV-TS, a time-series RNN-based model, manages to achieve robust performance and efficiency at the same time. In contrast, other RNN models are generally perceived to perform inadequately in both aspects for time-series tasks. Moreover, it holds a comparative advantage over other types of models. In comparison to MLP-based models, RWKV-TS exhibits better performance, while compared to Transformer-based models, it demonstrates enhanced efficiency.

Here we summarize our key contributions as follows:

\begin{itemize}
    \item We propose RWKV-TS, which is a RNN model with  $O(L)$ time complexity and memory usage. 
    \item The newly suggested RWKV-TS achieves performance similar to the current top-performing methods while notably cutting down on latency and memory usage.
    \item The success of RWKV-TS demonstrates significant improvements over existing RNN models, emphasizing the considerable potential of RNNs in the time series tasks. 
\end{itemize}

The remaining structure of this paper is as follows: Section 2 provides a brief summary of related works. Section 3 details the proposed model architecture. In Section 4, we conduct a comprehensive and in-depth evaluation of time-series analysis performance across six major time-series analysis tasks by comparing our proposed method with various state-of-the-art baseline models. Finally, Section 5 concludes our findings. Due to space constraints, a more extensive review of related works, experimental results, and theoretical analyses will be provided in the appendix.

%% file: sections/Related_Work.tex
Recent advancements have witnessed the ascendancy of Transformer models in time series tasks owing to their parallelization capabilities and attention mechanisms, enabling efficient processing of sequential data. Similarly, CNNs have demonstrated prowess in extracting local patterns and features from temporal data, exhibiting promising results in various time series domains.

\paragraph{RNN based models} Traditional Recurrent Neural Networks (RNNs) have long been the preferred choice for time series forecasting tasks due to their ability to handle sequential data. Continuous efforts have been made to employ RNNs for short-term and probabilistic predictions, resulting in significant advancements\cite{lstnet,MQRNN,RNN-Adap,C2FAR}. However, in the domain of Long-Term Sequence Forecasting (LTSF) with extended historical windows and prediction horizons, RNNs are considered ineffective in capturing long-term dependencies, leading to their gradual abandonment\cite{zhou2021informer,zhou2022fedformer}. 
The emergence of RWKV-TS aims to challenge and change this situation, attempting to address these limitations.
On the other hand, novel RNN architectures have achieved significant success and attention in the domain of large language models. 
Large-scale RNN models like RWKV\cite{Peng2023RWKVRR} and RNN variants such as Retentive Network\cite{Sun2023RetentiveNA} and Mamba\cite{Gu2023MambaLS} have demonstrated performance comparable to Transformer models, while also exhibiting superior efficiency.

\paragraph{Transformer based models} In recent years, there has been a considerable amount of research attempting to apply Transformer models to long-term time series prediction. Here, we summarize some of these works. The LogTrans model\cite{Log-transformer-shiyang-2019} utilizes convolutional self-attention layers with a LogSparse design to capture local information and reduce spatial complexity. The Informer\cite{zhou2021informer} introduces a ProbSparse self-attention mechanism with distillation techniques to efficiently extract the most critical information. Drawing inspiration from traditional time series analysis methods, the Autoformer\cite{wu2021autoformer} incorporates decomposition and autocorrelation concepts. The FEDformer\cite{zhou2022fedformer} employs Fourier-enhanced structures to achieve linear complexity. Additionally, the Pyraformer\cite{pyraformer} applies pyramid attention modules, encompassing inter-scale and intra-scale connections, also achieving linear complexity.  Additionally, it is worth mentioning the direct use of pre-trained GPT models\cite{Zhou2023OneFA} for time-series tasks. 

\paragraph{MLP based models} Multilayer Perceptrons (MLPs) are widely employed in time series forecasting\cite{n-beats,nhits}. Recently, DLinear gained an edge over the state-of-the-art Transformer-based models by simply adding a linear layer and a channel-agnostic strategy\cite{dlinear}. The success of DLinear propelled the development of numerous MLP models in LTSF, including MTS-Mixers\cite{Mts-mixers}, TSMixer\cite{TSMixer}, and TiDE\cite{tide}. The accomplishments of these MLP-based models have raised questions about the necessity of employing complex and intricate Transformers for time series forecasting. 

\paragraph{CNN based models}  CNNs have shown significant performance in the field of time series \cite{tcn,tcn2,MLCNN}. Recently, models based on CNNs such as MICN\cite{micn}, TimesNet\cite{wu2023timesnet}, and SCINet\cite{SCInet} have exhibited impressive results in the LTSF domain.

%% file: sections/Model_Architecture.tex
\begin{figure*}[htbp]
    \centering
    \includegraphics[width=0.9\textwidth]{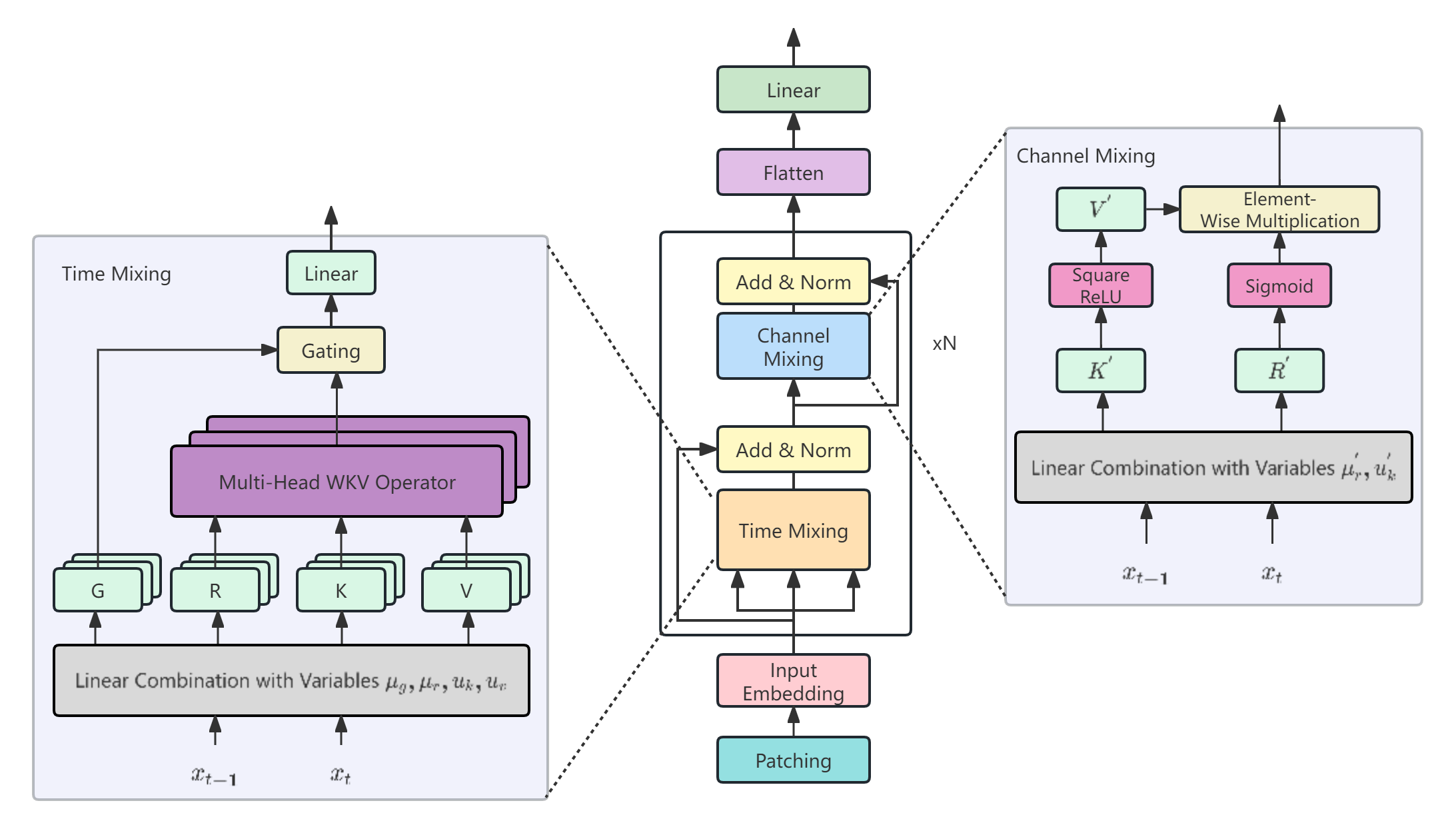} 
    \caption{Architecture of the RWKV-TS. RWKV-TS comprises three main components: an input module, RWKV backbone, and an output module. Firstly, the input module applies instance normalization to each channel's univariate series and segments them into patches. These patches serve as input tokens for RWKV-TS. Then, the input tokens proceed into the RWKV backbone, which comprises Time-mixing and Channel-mixing modules. Finally, the output of the last layer of the RWKV backbone is flattened and projected to predict the target.}
    \label{fig:model_arch}
\end{figure*}

Our RWKV-TS, depicted in Figure \ref{fig:model_arch}, utilizes the RWKV encoder as its backbone architecture. We address the following problem: given a dataset containing multiple multivariate time series $x \in R^{L \times M}$, each with a input sequence length $L$ and $M$ univariate. $x_i \in R^{L}$ presents a univariate instance and $i$ is one of $M$. Next, we'll outline the sequential order of modules through which the time series data will pass.

\subsection{Normalization and Pathcing}
\paragraph{Instance Normalization.} The first step is instance normalization. This technique has recently been proposed to mitigate the distribution shift effect between training and testing data \cite{kim2022reversible}. For each univariate time series instance, it was simply normalized with zero mean and unit standard deviation. Essentially, we normalize each instance before patching, and then the mean and deviation are reintroduced to the output prediction.

\paragraph{Patching.} In the second step, each input univariate time series $x_{i}$ is tokenized by patching\cite{Patchformer}. The univariate time series is segmented into patches, which can either overlap or not. If we denote the patch length as $P$ and the stride (the non-overlapping region between two consecutive patches) as $S$, then the patching process will generate a sequence of patches $x^p_{i} \in R^{N \times P}$, where $N$ is the number of patches and can be calculated as follow:
\begin{equation}
    N = \lfloor \frac{(L-P)}{S} \rfloor + 2
\end{equation}
By using patching, the number of input tokens was reduced from $L$ to $N$. This reduction helps the model to utilize fewer computations and less memory. Finally, the sequence of patches $x^p_{i} \in R^{N \times P}$was projected to the input tokens $x^t_{i} \in R^{N \times D}$ through a learnable projection matrix, where $D$ is the model dimension of the RWKV backbone.

\subsection{RWKV Blocks}

The RWKV backbone is structured using stacked residual blocks, with each block containing a time-mixing and a channel-mixing sub-block. These components embody recurrent structures designed to leverage past information. RWKV features both parallel and recurrent modes. The parallel mode is more advantageous for parallel training and offers higher computational efficiency, serving as the primary mode for code implementation. The main formulas for the parallel mode are presented below. The recurrent mode will be provided at the end of this subsection.

\subsubsection{Time Mixing Sub-block}
\paragraph{Token Shift}

In RWKV-TS, trainable variable $\mu_g$, $\mu_r$, $\mu_k$, $\mu_v$ are used in a linear combination of $x_t$ and $x_{t-1}$, to achieve a simple time mixing, which interpolate between the inputs of the current and previous time-steps. The combination of shifted previous step and current step was linear projected through projection matrix within the block:

\begin{align}
\label{eq:time-mix0}
g_t &= W_g \cdot (\mu_g \odot x_t + (1 - \mu_g) \odot x_{t-1} ), \\
\label{eq:time-mix1}
r_t &= W_r \cdot (\mu_r \odot x_t + (1 - \mu_r) \odot x_{t-1} ), \\
\label{eq:time-mix2}
k_t &= W_k \cdot (\mu_k \odot x_t + (1 - \mu_k) \odot x_{t-1} ), \\
\label{eq:time-mix3}
v_t &= W_v \cdot (\mu_v \odot x_t + (1 - \mu_v) \odot x_{t-1} ),
\end{align}

Token shift is a simple and intuitive way of "time mixing", ensuring effective information transmission in terms of its mechanism. Furthermore, these operations are all linear, allowing for easy parallel computation.

\paragraph{Multi-head WKV Operator}

In RWKV-TS, the WKV operator mirrors to self-attention but with a linear time and space complexity. This recurrent behavior in RWKV-TS is articulated through the time-dependent update of the WKV vectors.  The formula of single head WKV operator is given by:
\begin{equation}  
    \label{eq:wkv}
    wkv_{t} = \mathrm{diag}(u)\cdot k_{t}^\mathrm{T} \cdot v_{t} + \sum_{i=1}^{t-1} \mathrm{diag}(w)^{t-1-i} \cdot  k_{i}^\mathrm{T} \cdot v_{i} \\
\end{equation}
where $w$ and $u$ are two trainable parameters. The parameter $u$ is a bonus that rewards the model for encountering a token for the first time, specifically the current token. This helps the model pay more attention to the current token and circumvents any potential degradation of $w$. Another important parameter is $w$, which is a channel-wise time decay vector per head. Furthermore, we transform parameter $w$ as follows:
\begin{equation}
    w = \exp(-\exp(w))  
\end{equation}
This transformation ensures that all values of ${w}$ are within the range $(0,1)$, ensuring that $\mathrm{diag}(w)$ represents a contraction matrix.

Unlike the single head used in vanilla RWKV, RWKV-TS introduces a multi-head mechanism to enhance the model's capacity. Multi-head WKV was formally described by the following equation:
\begin{equation}
    \label{eq:multi-head_wkv}
    multihead\; wkv_t = Concat(wkv^1_t, ..., wkv^h_t)
\end{equation}
Where $h$ is the number of heads. However, in practice, we do not directly perform concatenation. Instead, we cleverly employ reshape operations for parallel computation, and finally reshape the result to an equivalent form. This explanation is provided for ease of understanding.

\paragraph{Output Gating}

Output gating is implemented using SiLU and receptance. The output vector $o_t$ per head is given by:
\begin{equation}
    o_t = ( SiLU(g_t) \odot \mathrm{LayerNorm}(r_t \cdot wkv_t) )W_o
\end{equation}
Where LayerNorm operates on each of $h$ heads separately, which is also equivalent to the GroupNorm\cite{Wu2018GroupN} operation on $h$ groups.
\subsubsection{Channel Mixing Sub-block}
In the channel-mixing block, channels are mixed by strong non-linear operations as follow:
\begin{align}
\label{eq:channel-mix0}
k^{'}_t &= W^{'}_g \cdot (\mu^{'}_k \odot x_t + (1 - \mu^{'}_k) \odot x_{t-1} ), \\
\label{eq:channel-mix1}
r^{'}_t &= W^{'}_r \cdot (\mu^{'}_r \odot x_t + (1 - \mu^{'}_r) \odot x_{t-1} ), \\
v'_t &= \mathrm{ReLU}^2(k'_t) \cdot W^{'}_{v} \\
o'_t &= \mathrm{Sigmoid}(r'_t) \odot v'_t
\end{align}
where we adopt the squared ReLU activation function to enhance the non-linearity.

\subsubsection{Recurrent Mode}
Readers might be puzzled, as the formulas presented earlier do not immediately reveal that RWKV is an RNN. \textbf{Why is RWKV referred to as an RNN?} A significant feature of RWKV is that it has one model with two sets of formulas, allowing the parallel mode to be rewritten as the recurrent mode, and these two modes are entirely equivalent. The \textit{\textbf{wkv}} term can alternatively be written in a recurrent mode:
\begin{align}
    \label{eq:wkv_recurrent}
    wkv_{t} &= s_{t-1} + \mathrm{diag}(u)\cdot k_{t}^\mathrm{T} \cdot v_{t} \\
    s_t &= \mathrm{diag}(w) \cdot s_{t-1} + k_{t}^\mathrm{T} \cdot v_{t}
\end{align} 
where $s_t$ is state of current time step, and $s_{t-1}$ is the state from last time step.

\subsection{Flatten and Loss Function}

\paragraph{Flatten.}  Finally, a flatten layer with a linear head is employed to obtain the prediction result $y_i$

\paragraph{Loss Function.} The loss function used for our model is the mean squared error (MSE), which is defined as:
\begin{equation}
\text{MSE}(y, \hat{y}) = \frac{1}{n} \sum_{i=1}^{n} (y_i - \hat{y}_i)^2\
\end{equation}

\subsection{Complexity Analysis}

We conduct a complexity analysis of RWKV-TS as shown in Table \ref{tab:complexity}. As RWKV-TS is a linear RNN model, both its time and space complexities are $O(L)$, which is superior to the $O(L^2)$ complexity of the vanilla Transformer and comparable to the most efficient DLinear model. Additionally, because RWKV-TS follows an encoder-only architecture, its test step is also 1. This ensures both efficiency and avoids the problem of error accumulation seen in traditional RNN models. 

\input{tables/complexity}

\subsection{Advantage over Traditional RNN Models}
Compared to traditional RNNs, RWKV-TS has the following significant advantages:

\paragraph{Parallel Computation:} RWKV-TS uses a linear RNN design, which allows for parallel computation, resulting in high computational efficiency and the ability to scale up. In contrast, LSTM/GRU cannot be parallelized due to their non-linear dependence on the last hidden state, leading to lower efficiency and an inability to scale up\cite{Kaplan2020ScalingLF}.

\paragraph{Enhanced Long-Range Information Capturing Capability:} Traditional RNNs such as LSTM and GRU encounter a noticeable bottleneck when the sequence length exceeds 100, where the token loss cannot be further reduced\cite{Kaplan2020ScalingLF}; whereas RWKV has been demonstrated to maintain strong information capturing ability even when the sequence length reaches 4096, with token loss continuously decreasing\cite{Peng2023RWKVRR}. This is largely due to the well-designed token shift and time decay mechanisms in RWKV-TS, which facilitate the easier transmission of information within the model.

\paragraph{Encoder-only Architecture:} RWKV-TS incorporates the most advanced ideas\cite{Patchformer,zhou2021informer} in the field of time series, using an encoder-only architecture to process time series information, abandoning the traditional RNN's recurrent iteration pattern. This allows RWKV-TS to have only one inference step and eliminates error accumulation.

%% file: tables/complexity.tex
\begin{table}[h]
\vspace{-0.1cm}
        \centering
        \scalebox{0.9}
        % \scriptsize
        {
        	{%
                \begin{tabular}{c|ccc} \hline
                Method      & Time & Memory & Test Step \\\hline
                RWKV-TS  & $O(L)$   & $O(L)$    & 1  \\
                LSTM& $O(L)$& $O(L)$&$L$\\\hline
                DLinear  & $O(L)$   & $O(L)$    & 1  \\\hline
                TimesNet  & $O(k^2L)$   & $O(L)$    & 1  \\\hline
                Transformer & $O(L^2)$  & $O(L^2)$     & $L$\\
                Informer    & $O(LlogL)$   & $O(LlogL)$     & 1    \\
                Autoformer  & $O(LlogL)$   & $O(LlogL)$     & 1   \\
                LogTrans& $O(LlogL)$&  $O(L^2)$& 1  \\
                PatchTST& $O(L^2)$& $O(L^2)$& 1\\\hline               
                \end{tabular}%
        	}
            }
            \vspace{-0.2cm}
        \caption{Time Complexity and Space Complexity Comparison. 
        $L$ is the sequence length, $k$ is the kernel size of convolutions. 
        Models are categorized into four types based on their architecture: RNN, MLP, CNN, and Transformer.}
\label{tab:complexity}
\vspace{-0.3cm}
\end{table}

%% file: sections/Experiment.tex
Extensive experiments on major types of downstream tasks validates the performance and efficiency of RWKV-TS. Downstream tasks include time-series classification, anomaly detection, imputation, short-term forecasting, long-term forecasting, and few-shot prediction. 

\textbf{Baseline Models} We selected representative baseline models and referenced their results from \cite{timesnet}, which includes most recent and fairly extensive empirical studies on time-series. The baseline models include 

\begin{itemize}
    \item CNNs-based models: TimesNet\cite{timesnet};
    \item MLPs-based models: LightTS\cite{lightts} and DLinear\cite{dlinear};
    \item Transformers-based models: Informer\cite{zhou2021informer}, Autoformer\cite{wu2021autoformer}, FEDformer\cite{zhou2022fedformer}, Non-stationary Transformer\cite{non-stationary}, ETSformer\cite{woo2022etsformer}, PatchTST\cite{Patchformer}.
\end{itemize}

Additionally, for short-term forecasting, we employed N-HiTS\cite{nhits} and N-BEATS\cite{n-beats}. Anomaly detection utilized Anomaly Transformer\cite{xu2021anomaly}. For the classification task, we opted for four categories of baseline models: classical methods\cite{xgboost,ROCKET}, RNN-based models\cite{lstnet,lssl,lstm}, CNN-based models\cite{tcn,timesnet}, and Transformer-based models\cite{pyraformer}.
The pre-trained models\cite{Zhou2023OneFA} were not considered as baselines as it would be an unfair comparison. Both the RWKV-TS and baseline models in this paper were trained from scratch.

For more experimental details, please refer to the appendix \ref{appendix:Experimental details}.

\subsection{\textbf{Efficiency Analysis}}
Firstly, for RWKV-TS, its most direct advantage over PatchTST (a transformer-based model) and TimesNet (a CNN-based model) arises from its $O(L)$ time and space complexity, granting RWKV-TS an efficiency edge.

Analysis of computational costs is crucial for examining the practicality of RWKV-TS. The outcomes are presented in Table \ref{tab:cost}. Each baseline model and RWKV-TS are configured with hidden dimensions of 768 and comprised of 3 layers. Computational costs were evaluated using a batch from ETTh2 (with a batch size of 128).

The outcomes illustrate that RWKV-TS significantly improves time efficiency and reduces parameter count when compared to baselines with similar model dimensions. FEDformer, TimesNet, and PatchTST are strong baselines, each being state-of-the-art models in their respective domains; RWKV-TS exhibits competitive performance against them in various domains, but notably outperforms them in terms of efficiency.
For a more comprehensive comparison of efficiency, additional details can be found in the appendix \ref{appendix:efficiency}.

\input{tables/cost}

\subsection{\textbf{Long-term Forecasting }}

\input{tables/Long-term}

For long-term forecasting, the experiment was conducted as follows: We utilized eight well-known real-world benchmark datasets, comprising Weather, Traffic, Electricity, ILI, and 4 ETT datasets (ETTh1, ETTh2, ETTm1, ETTm2), for evaluating long-term forecasting performance.

The experimental outcomes are presented in Table \ref{tab:100_percent_ett_main}. RWKV-TS demonstrates a performance level comparable to PatchTST and surpasses other baseline models. Particularly noteworthy is its out-performance compared to the recent state-of-the-art method, TimesNet, demonstrate a relative 12.58 \% average MSE reduction and 4.38\% average MAE reduction.

\subsection{\textbf{Short-term Forecasting }}

\input{tables/short_term}

To comprehensively assess diverse algorithms in forecasting tasks, we additionally perform experiments on short-term forecasting  using the M4 dataset \cite{makridakis2018m4}, which encompasses marketing data of various frequencies.

The results displayed in Table \ref{tab:short_term} indicate that RWKV-TS has outperformed several transformer-based models such as ETSformer, FEDformer, Informer, Autoformer, and Non-stationary Transformer. 
Additionally, it also indicates that compared to MLP-based models like DLinear and LightTS, RWKV-TS exhibits better performance.
The difference between RWKV-TS and state-of-the-art models like TimesNet and N-BEATS is also marginal.

\subsection{\textbf{Few-shot Forecasting }}

\input{tables/few_shot_main}

To comprehensively assess the representation capacity of RWKV-TS for time series analysis, we conducted experiments within the few-shot learning paradigm.

Following conventional experimental settings, each time series is partitioned into three segments: training data, validation data, and test data. For the few-shot learning approach, only 10\% of the time-steps from the training data are utilized.

The outcomes of the 10\% few-shot learning experiments are presented in Table \ref{tab:few_shot_main}. RWKV-TS outperforms TimesNet, DLinear, PatchTST, and other methodologies, demonstrating superior performance. Conventionally, CNN-based and single MLP-based models are regarded as more adept at utilizing data for training and are suitable for few-shot learning techniques. In contrast to the convolution-based TimesNet and MLP-based DLinear models, RWKV-TS showcases a relative average MSE reduction of 28.95\% and 7.92\% respectively.

\subsection{\textbf{Time Series Classification}}

\input{tables/classification}

To assess the model's aptitude for high-level representation learning, we adopt sequence-level classification. Following the same protocol as TimesNet, we select 10 multivariate UEA classification datasets \cite{UEA} for evaluation, encompassing gesture recognition, action identification, audio recognition, medical diagnosis, and other practical tasks.

The outcomes depicted in Table \ref{tab:classification} reveal that RWKV-TS achieves an average accuracy of 73.10\%, outperforming most baselines, with the exception of TimesNet (73.60\%). RWKV-TS not only outperforms all other RNN models but also exhibits superiority over other transformer-based and MLP-based models. This substantiates that linear RNNs can indeed produce robust time series representations.

\subsection{\textbf{Time Series Anomaly Detection}}

\input{tables/anomaly}

Detecting anomalies in time series is crucial across various industrial applications, spanning from health monitoring to space and earth exploration. We compare models using five frequently utilized datasets, such as SMD\cite{SMD}, MSL\cite{MSL_SMAP}, SMAP\cite{MSL_SMAP}, SWaT\cite{SWaT}, and PSM\cite{PSM}. To ensure a fair comparison, all baseline models exclusively use the classical reconstruction error, aligning with the setting employed in TimesNet.

As shown in Table \ref{tab:anomaly}, RWKV-TS also showcases good performance, achieving an averaged F1-score of 83.89\%, which is close to the performance of other state-of-the-art (SOTA) models. This suggests that RWKV-TS is proficient in detecting infrequent anomalies within time series, thus illustrating its versatility as a multi-functional time series model.

\subsection{\textbf{Imputation}}

\input{tables/imputation}

We conducted experiments on six widely used real-world datasets: 4 ETT datasets (ETTh1, ETTh2, ETTm1, ETTm2), Electricity, and Weather, which commonly contain missing data. Following the settings adopted in TimesNet, various random mask ratios (12.5\%, 25\%, 37.5\%, 50\%) of time points were selected to evaluate different proportions of missing data.

The results in Table \ref{tab:imputation} reveal that while RWKV-TS performs less effectively than state-of-the-art models such as TimesNet and PatchTST, it still outperforms MLP-based models and certain Transformer-based models. RWKV-TS is a unidirectional RNN model, capable of only observing information preceding the missing values. In contrast, Transformer-based models are bidirectional models, able to capture information both before and after missing values, hence exhibiting better performance. Future work might transform RWKV-TS into a bidirectional model, maybe Bi-RWKV-TS, which could potentially lead to improvements in its performance on this task.

%% file: tables/cost.tex
\begin{table}[h]
\vskip -0.1in
%\vskip 0.15in
\begin{center}
\begin{small}
\scalebox{0.9}{
\setlength\tabcolsep{3pt}
\begin{tabular}{c|ccc}
\toprule
Model& Parameters& Training Time(s) & Inference Time(s)\\
\midrule
RWKV-TS-768 &24M &\textbf{0.067}&\textbf{0.018} \\
FEDformer-768&33M &0.208&0.056  \\
TimesNet-768&42M &5.723&2.162 \\
PatchTST-768 &20M &0.457&0.123 \\
\bottomrule
\end{tabular}
}
\end{small}
\end{center}
\captionsetup{font=small} 
\caption{Comparison of Training and Inference Costs: The Training Time is measured per step, while Inference Time is measured per batch.}
\label{tab:cost}
\vskip -0.2in

\end{table}

%% file: tables/Long-term.tex
\begin{table}[htbp]
%\vskip 0.15in
\begin{center}
\begin{small}
    \scalebox{0.7}{
    \setlength\tabcolsep{3pt}
    \begin{tabular}{c|cc|cc|cc|cc|cc}
    \toprule
    
    \multirow{2}{*}{Methods} 
    &\multicolumn{2}{c|}{RWKV-TS}  & \multicolumn{2}{c|}{TimesNet}&\multicolumn{2}{c|}{ETSformer}&\multicolumn{2}{c|}{LightTS}&\multicolumn{2}{c}{DLinear}\\
    &MSE&MAE&MSE&MAE&MSE&MAE&MSE&MAE&MSE&MAE \\
    
    \midrule
    
    Weather &\color{red}\textbf{0.231} & \color{red}\textbf{0.266} & 0.259& 0.287& 0.271& 0.334& 0.261& 0.312&0.249 & 0.300 \\
    ETTh1 & 0.433& 0.445& 0.458& 0.450& 0.542& 0.510& 0.491& 0.479&\color{red}\textbf{0.423} &\color{red}\textbf{0.437} \\
    ETTh2 & \color{red}\textbf{0.375}& \color{red}\textbf{0.412}& 0.414& 0.427& 0.439& 0.452& 0.602& 0.543&0.431 & 0.447\\
    ETTm1 &0.376 & 0.401& 0.400 &0.406 &0.429 &0.425 &0.435 &0.437 &\color{red}\textbf{0.357} &\textbf{0.378} \\
    ETTm2 &0.287 & 0.338& 0.291& 0.333& 0.293& 0.342& 0.409& 0.436& \color{red}\textbf{0.267}& \color{red}\textbf{0.334}\\
    ILI &\color{red}\textbf{1.910}&\color{red}\textbf{0.925}& 2.139& 0.931& 2.497& 1.004& 7.382& 2.003& 2.169&1.041 \\
    ECL &\textbf{0.159} &\textbf{0.253} & 0.192& 0.295& 0.208& 0.323& 0.229& 0.329&0.166 &0.263 \\
    Traffic &\color{red}\textbf{0.398} & \color{red}\textbf{0.276} & 0.620& 0.336& 0.621& 0.396& 0.622& 0.392& 0.434& 0.295\\
    \midrule
    
    Average & \color{red}\textbf{0.521} & \color{red}\textbf{0.414} & 0.596 & 0.433 & 0.662 & 0.473 & 1.303 & 0.616 & 0.562 & 0.436\\

    \midrule
    \multirow{2}{*}{Methods} 
    &\multicolumn{2}{c|}{FEDformer}&\multicolumn{2}{c|}{PatchTST}&\multicolumn{2}{c|}{Stationary}&\multicolumn{2}{c|}{Autoformer}&\multicolumn{2}{c}{Informer} \\
    &MSE&MAE&MSE&MAE&MSE&MAE&MSE&MAE&MSE&MAE \\
    \midrule
    Weather & 0.309& 0.360&\textbf{0.225} &\textbf{0.264} &0.288& 0.314& 0.338& 0.382& 0.634& 0.548\\
    ETTh1 & 0.440& 0.460&\textbf{0.413}&\textbf{0.430} & 0.570& 0.537& 0.496& 0.487& 1.040& 0.795\\
    ETTh2 & 0.437& 0.449&\textbf{0.330} &\textbf{0.379} &0.526& 0.516& 0.450& 0.459& 4.431& 1.729\\
    ETTm1 & 0.448 &0.452 & \textbf{0.351}&\color{red}\textbf{0.387} &0.481 &0.456& 0.588 &0.517 &0.961 &0.734\\
    ETTm2 & 0.305& 0.349&\textbf{0.255} & \textbf{0.315}& 0.306& 0.347& 0.327& 0.371& 1.410& 0.810\\
    ILI & 2.847& 1.144&\textbf{1.443} &\textbf{0.798} &2.077& 0.914& 3.006& 1.161& 5.137& 1.544\\
    ECL & 0.214& 0.327&0.161 &0.253 &0.193& 0.296& 0.227& 0.338& 0.311& 0.397\\
    Traffic & 0.610& 0.376&\textbf{0.390} &\textbf{0.264} &0.624& 0.340& 0.628& 0.379& 0.764& 0.416\\
    \midrule
    
    Average & 0.701 & 0.489 & \textbf{0.446} & \textbf{0.386} & 0.633 & 0.465 & 0.757 & 0.511 & 1.836 & 0.871 \\
    
    \bottomrule
    
    \end{tabular}
}
\end{small}
\end{center}
\captionsetup{font=small} 
\caption{Long-term forecasting task. All the results are averaged from 4 different prediction lengths, that is \{24, 36, 48, 60\} for ILI and \{96, 192, 336, 720\} for the others. Bold black is the best, red is the second best.}

\label{tab:100_percent_ett_main}
\end{table}

%% file: tables/short_term.tex
\begin{table}[htbp]
\vskip -0.10in
%\vskip 0.15in
\begin{center}
\begin{small}
\scalebox{0.7}{
    \setlength\tabcolsep{3pt}
    \begin{tabular}{c|cccccc}
    \toprule
    
    Methods&RWKV-TS&TimesNet&PatchTST&N-HiTS&N-BEATS& ETSformer \\    
    \midrule
    
    SMAPE &12.021 & \textbf{11.829}&12.059& 11.927& 11.851& 14.718 \\
    MASE & 1.631 & \textbf{1.585}&1.623 & 1.613 & 1.599 &2.408 \\
    OWA &0.870 & \textbf{0.851}&0.869 &0.861 &0.855 &1.172 \\
    \midrule
    Methods&LightTS& DLinear &FEDformer &Stationary&Autoformer&Informer \\
    \midrule
    SMAPE &13.525& 13.639 &12.840 &12.780 &12.909 &14.086\\
    MASE &2.111 &2.095 &1.701 &1.756 &1.771  &2.718\\
    OWA &1.051 &1.051 &0.918 &0.930 &0.939 & 1.230\\
    \bottomrule
    \end{tabular}
}
\end{small}
\end{center}
\vskip -0.1in
\captionsetup{font=small} 
\caption{Short-term forecasting task on M4. The prediction lengths are in [6, 48] and results are weighted averaged from several datasets under different sample intervals.}
\label{tab:short_term}
\end{table}

%% file: tables/few_shot_main.tex
\begin{table}[h]
% \vskip 0.15in
\begin{center}
\begin{small}
\scalebox{0.65}{
\setlength\tabcolsep{3pt}
\begin{tabular}{c|cc|cc|cc|cc|cc}
\toprule

\multirow{2}{*}{Methods} 
&\multicolumn{2}{c|}{RWKV-TS}  & \multicolumn{2}{c|}{TimesNet}&\multicolumn{2}{c|}{DLinear}&\multicolumn{2}{c|}{FEDformer}&\multicolumn{2}{c}{PatchTST} \\
&MSE&MAE&MSE&MAE&MSE&MAE&MSE&MAE&MSE&MAE \\

\midrule

Weather &0.232&0.268&0.279&0.301&0.301&0.283&0.284&0.324&0.241&0.279\\
ETTh1  &0.616&0.542&0.869&0.628&0.691&0.599&0.638&0.561&0.633&0.542\\
ETTh2  &0.411&0.425&0.479&0.465&0.608&0.538&0.466&0.475&0.415&0.431 \\
ETTm1  &0.561&0.486&0.676&0.537&0.411&0.429&0.721&0.605&0.501&0.466\\
ETTm2 &0.325&0.357&0.319&0.353&0.316&0.368&0.463&0.488&0.296&0.343\\
ECL  &0.180&0.274&0.323&0.392&0.180&0.280&0.346&0.428&0.180&0.269\\
Traffic  &0.441&0.314&0.951&0.535&0.496&0.371&0.663&0.425&0.430&0.305\\

\midrule
Average & \color{red}\textbf{0.395} & \color{red}\textbf{0.380} & 0.556 & 0.458 & 0.429 & 0.409 & 0.511 & 0.472 & \textbf{0.385} & \textbf{0.376} \\

\midrule
\multirow{2}{*}{Methods} 
&\multicolumn{2}{c|}{Autoformer}&\multicolumn{2}{c|}{Stationary}&\multicolumn{2}{c|}{ETSformer}&\multicolumn{2}{c|}{LightTS}&\multicolumn{2}{c}{Informer} \\
&MSE&MAE&MSE&MAE&MSE&MAE&MSE&MAE&MSE&MAE \\
\midrule

Weather &0.300&0.342&0.318&0.322& 0.317 & 0.359 & 0.289 & 0.322 & 0.597 & 0.494\\
ETTh1  &0.701&0.596&0.914&0.639& 1.179 & 0.833 & 1.375 & 0.877 & 1.199 & 0.808\\
ETTh2  &0.488&0.499&0.461&0.454& 0.893 & 0.713 & 2.655 & 1.159 & 3.871 & 1.512 \\
ETTm1  &0.802&0.628&0.797&0.577& 0.979 & 0.714 & 0.970 & 0.704 & 1.192 & 0.820\\
ETTm2 &1.341&0.930&0.332&0.366& 0.447 & 0.487 & 0.987 & 0.755 & 3.369 & 1.439  \\
ECL  &0.431&0.478&0.443&0.479& 0.659 & 0.617 & 0.441 & 0.488 & 1.194 & 0.890 \\
Traffic  &0.749&0.446&1.453&0.815& 1.913 & 0.936 & 1.247 & 0.684 & 1.534 & 0.811 \\

\midrule

Average & 0.687 & 0.559 & 0.674 & 0.522 & 0.912 & 0.665 & 1.137 & 0.712 & 1.850 & 0.967  \\

\bottomrule

\end{tabular}
}
\end{small}
\end{center}
\captionsetup{font=small} 
\caption{Few-shot learning task on 10\% data. All the results are averaged from 4 different prediction lengths (\{96, 192, 336, 720\}). 
Bold black is the best and red is the second best.}
\label{tab:few_shot_main}
\end{table}

%% file: tables/classification.tex
\begin{table}[htbp]
\begin{center}
\begin{small}
\scalebox{0.65}{
    \setlength\tabcolsep{3pt}
    \begin{tabular}{c|cccc|cc|cc}
    \hline
    
    \multirow{2}{*}{Methods} & \multicolumn{4}{c|}{RNNs} &
    \multicolumn{2}{c|}{Classical methods} & \multicolumn{2}{c}{MLPs} \\
    &RWKV-TS&LSTM&LSTNet& LSSL&XGBoost&Rocket&DLinear &LightTS\\
    
    \hline
    EthanolConcentration&34.6&32.3&43.7 &45.2 &39.9 &31.1&32.6 &29.7\\
    FaceDetection&67.3&57.7&63.3 &64.7 &65.7 &66.7&68.0 &67.5\\
    Handwriting&34.6&15.2&15.8 &58.8 &25.8 &24.6&27.0 &26.1\\
    Heartbeat&77.1&72.2&73.2 &75.6 &77.1 &72.7&75.1 &75.1\\
    JapaneseVowels&98.4&79.7&86.5 &96.2 &98.1 &98.4 &96.2 &96.2\\
    PEMS-SF&83.8&39.9&98.3 &75.1 &86.7 &86.1 &75.1 &88.4 \\
    SelfRegulationSCP1&91.8&68.9&84.6 &90.8 &84.0 &90.8 &87.3 &89.8 \\
    SelfRegulationSCP2&57.2&46.6&48.9 &53.3 &52.8 &52.2 &50.5 &51.1 \\
    SpokenArabicDigits&98.9&31.9&69.6 &71.2 &100.0 &100.0&81.4 &100.0\\
    UWaveGestureLibrary&86.9&41.2&75.9 &94.4 &87.8 &85.9&82.1 &80.3 \\
    \hline
    Average &\color{red}\textbf{73.1}&48.6&66.0 &72.5 &71.8 &70.9 &67.5 &70.4\\
    \hline
    
    \end{tabular}
}
\scalebox{0.65}{
    \setlength\tabcolsep{3pt}
    \begin{tabular}{c|cccccc|cc}
    \hline
    
    \multirow{2}{*}{Methods} & \multicolumn{6}{c|}{Transformers} & \multicolumn{2}{c}{CNNs} \\
    &Transformer& Informer& Pyraformer& Auto.& FED.& ETS. & TimesNet & TCN\\
    
    \hline
    EthanolConcentration&32.7&31.6 &30.8 &31.6 &31.2 &28.1 &35.7&28.9\\
    FaceDetection&67.3 &67.0 &65.7 &68.4 &66.3 &68.6&52.8\\
    Handwriting&32.0&32.8 &29.4 &36.7 &28.0 &32.5 &32.1&53.3 \\
    Heartbeat&76.1&80.5 &75.6 &74.6 &73.7 &71.2 &78.0 &75.6 \\
    JapaneseVowels&98.7&98.9 &98.4 &96.2  &98.4 &95.9 &98.4 &98.9 \\
    PEMS-SF&82.1&81.5 &83.2 &82.7 &80.9 &86.0 &89.6&68.8 \\
    SelfRegulationSCP1&92.2&90.1 &88.1 &84.0 &88.7 &89.6&91.8&84.6\\
    SelfRegulationSCP2&53.9&53.3 &53.3 &50.6 &54.4 &55.0&57.2&55.6 \\
    SpokenArabicDigits&98.4&100.0 &99.6 &100.0 &100.0 &100.0 &99.0&95.6 \\
    UWaveGestureLibrary&85.6&85.6 &83.4 &85.9 &85.3 &85.0 &85.3&88.4 \\
    \hline
    Average &71.9&72.1 &70.8 &71.1 &70.7 &71.0 &\textbf{73.6}&70.3 \\
    \hline
    
    \end{tabular}
}
\end{small}
\end{center}
\vskip -0.1in
\caption{Full results for the classification task.
Bold black is the best and red is the second best.}
\label{tab:classification}
\end{table}

%% file: tables/anomaly.tex
\begin{table}[htbp]
\vskip -0.1in
%\vskip 0.15in
\begin{center}
\begin{small}
\scalebox{0.72}
{
    \setlength\tabcolsep{3pt}
    \begin{tabular}{c|ccccccc}
    \toprule
    
    Methods& RWKV-TS & TimesNet & PatchTS. & ETS.& FED. & LightTS & DLinear \\

    \midrule
    SMD &84.33 &84.61 &84.62&83.13& 85.08& 82.53& 77.10  \\
    MSL &77.92&81.84&78.70&85.03&78.57&78.95&84.88 \\
    SMAP&68.71&69.39&68.82& 69.50& 70.76& 69.21& 69.26\\
    SWaT&91.40& 93.02&85.72 & 84.91& 93.19 &93.33& 87.52 \\
    PSM&97.10&  97.34&96.08& 91.76& 97.23& 97.15& 93.55 \\
    \midrule
    Average &83.89&85.24&82.79& 82.87& 84.97& 84.23& 82.46\\

    \midrule
    Methods&Stationary& Auto. & Pyra. & Anomaly & In.& LogTrans. & Trans.  \\
    \midrule
    SMD  &84.72 &85.11 &83.04& 85.49 &81.65 & 76.21& 79.56 \\
    MSL &77.50&79.05&84.86&83.31&84.06&79.57&78.68 \\
    SMAP& 71.09& 71.12& 71.09& 71.18& 69.92& 69.97& 69.70 \\
    SWaT& 79.88& 92.74& 91.78& 83.10& 81.43& 80.52& 80.37 \\
    PSM& 97.29& 93.29& 82.08& 79.40& 77.10 &76.74 &76.07 \\
    \midrule
    Average& 82.08& 84.26& 82.57 &80.50& 78.83& 76.60& 76.88\\
    \bottomrule
    \end{tabular}
}
\end{small}
\end{center}

\vskip -0.1in
\captionsetup{font=small} 
\caption{Anomaly detection task. We calculate the F1-score (as \%) for each dataset.}
\label{tab:anomaly}
\end{table}

%% file: tables/imputation.tex
\begin{table}[htbp]
%\vskip 0.15in
\begin{center}
\begin{small}
\scalebox{0.7}{
    \setlength\tabcolsep{3pt}
    \begin{tabular}{c|cc|cc|cc|cc|cc}
    \toprule
    
    \multirow{2}{*}{Methods} 
    &\multicolumn{2}{c|}{RWKV-TS} & \multicolumn{2}{c|}{TimesNet}&\multicolumn{2}{c|}
    {PatchTST}&\multicolumn{2}{c|}
    {ETSformer}&\multicolumn{2}{c}{LightTS} \\
    &MSE&MAE&MSE&MAE&MSE&MAE&MSE&MAE&MSE&MAE \\
    
    \midrule
    
    ETTm1& 0.149& 0.219&0.027 &0.107&0.047 &0.140 & 0.120& 0.253& 0.104& 0.218 \\
    ETTm2& 0.030& 0.111&0.022& 0.088&0.029 &0.102 & 0.208& 0.327& 0.046& 0.151\\
    ETTh1& 0.268& 0.310&0.078& 0.187&0.115 &0.224 & 0.202& 0.329& 0.284& 0.373\\
    ETTh2& 0.060& 0.164&0.049& 0.146&0.065 &0.163 & 0.367& 0.436& 0.119& 0.250\\
    ECL& 0.180& 0.293&0.092&0.210&0.072 &0.183 & 0.214& 0.339& 0.131& 0.262 \\
    Weather&0.038&0.066&0.030& 0.054&0.034 &0.055 & 0.076& 0.171& 0.055& 0.117\\
    
    \midrule
    
    Average & 0.120 & 0.193 & 0.049 & 0.132 & 0.060 & 0.144 & 0.197 & 0.309 & 0.123 & 0.228\\

    \midrule
    \multirow{2}{*}{Methods} 
    &\multicolumn{2}{c|}{DLinear}&\multicolumn{2}{c|}{FEDformer}&\multicolumn{2}{c|}{Stationary}&\multicolumn{2}{c|}{Autoformer}&\multicolumn{2}{c}{Informer} \\
    &MSE&MAE&MSE&MAE&MSE&MAE&MSE&MAE&MSE&MAE \\
    \midrule
    ETTm1& 0.093& 0.206& 0.062& 0.177& 0.036& 0.126&0.051& 0.150&  0.071& 0.188 \\
    ETTm2& 0.096& 0.208& 0.101& 0.215& 0.026& 0.099&0.029& 0.105& 0.156& 0.292\\
    ETTh1& 0.201& 0.306& 0.117& 0.246& 0.094& 0.201 &0.103& 0.214& 0.161& 0.279\\
    ETTh2& 0.142& 0.259& 0.163& 0.279& 0.053& 0.152& 0.055& 0.156& 0.337& 0.452\\
    ECL& 0.132& 0.260& 0.130& 0.259& 0.100& 0.218 &0.101& 0.225& 0.222& 0.328 \\
    Weather& 0.052& 0.110& 0.099& 0.203& 0.032& 0.059 &0.031& 0.057& 0.045& 0.104 \\
    \midrule
    Average& 0.119 & 0.224 & 0.112 & 0.229 & 0.056 & 0.142 & 0.061 & 0.151 & 0.165 & 0.273 \\
    
    \bottomrule
    
    \end{tabular}
}
\end{small}
\end{center}
\captionsetup{font=small} 
\caption{Imputation task. We randomly mask \{12.5\%, 25\%, 37.5\%, 50\%\} time points of 96-length time series. The results are averaged from 4 different mask ratios.}
\label{tab:imputation}
\end{table}

%% file: sections/Conclusion.tex
 Overall, RWKV-TS has achieved performance comparable to state-of-the-art (SOTA) across multiple time series tasks, with varying degrees of success against models like PatchTST and TimesNet on different tasks. The most notable advantage of RWKV-TS is its $O(L)$ time complexity and memory usage, which enable it to be efficient, fast, and have a small memory footprint. This makes RWKV highly competitive for deployment on end-devices with limited computational and memory resources.
Additionally, the success of RWKV-TS addresses the initial question of whether RNNs are truly no longer suitable for time series tasks. The answer might be No.
The empirical success of RWKV-TS highlights the resilience of RNNs in time series analysis. The model's ability to achieve competitive performance while addressing computational inefficiencies encourages further exploration and innovation in RNN-based approaches within the time series domain. Future research endeavors may delve deeper into enhancing the efficiency and effectiveness of RNN architectures tailored for diverse temporal data characteristics and applications.

%% file: sections/Appendix.tex
\appendix
\onecolumn

\section{Experimental details}
\label{appendix:Experimental details}

\subsection{Model Parameters}
By default, RWKV-TS includes 2 encoder layers with number of heads H = 2 and hidden dimension D = 128. The default model's parameter count is not fixed; it varies with the length of the input sequence and the prediction length.  Apart from the RWKV-TS utilized in the Efficiency experiment, which is configured with 3 layers and a hidden dimension of 768 to align with other architectures, all other instances of RWKV-TS across the experiments conform to the default model parameters.

\subsection{Hyperparameters}
We present the training hyperparameters here. By default, learning rate is 0.0001, learning rate schedule is cosine decay, no weight decay and the optimizer is AdamW optimizer.  Typically, datasets are trained for 10 epochs, and the training process employs an early stopping strategy.

\section{Experimental Results}

\subsection{Long-term Forecasting}
\label{appendix:Long-term Forecasting}

\paragraph{Datasets for Long-term Forecasting} We use 8 widely recognized multivariate datasets\cite{wu2021autoformer} for forecasting and representation learning purposes. The Weather dataset compiles 21 meteorological indicators in Germany, including humidity and air temperature. The Traffic dataset captures road occupancy rates from various sensors on San Francisco freeways. Electricity details the hourly electricity consumption of 321 customers. The ILI dataset gathers weekly data on the number of patients and the ratio of influenza-like illness. The ETT datasets, collected from two distinct electric transformers labeled as 1 and 2, each contain two different temporal resolutions: 15-minute intervals (m) and hourly intervals (h). Consequently, we have a total of four ETT datasets: ETTm1, ETTm2, ETTh1, and ETTh2.

In the table \ref{tab:100_percent_ett_full}, we present the detailed outcomes for Long-term Time-series Forecasting. RWKV-TS delivers performance on par with PatchTST and Dlinear, and significantly surpasses other benchmarks by a wide margin. RWKV-TS achieves the best performance on one dataset and ranks second-best on four datasets.

\input{tables/Long-term_full}

\subsection{Short-term Forecasting}
\label{appendix:Short-term Forecasting}

\paragraph{Datasets for Short-term Forecasting} The specifics of datasets for short-term forecasting are as follows: 
\begin{enumerate}
    \item M4 is a comprehensive and diverse dataset encompassing time series of various frequencies and sectors, including business, financial, and economic forecasting; 
    \item M3, while smaller than M4, also comprises time series from a range of domains and frequencies; 
    \item TOURISM is a dataset of tourism activities with varying frequencies and includes a significantly higher proportion of erratic series compared to M4; 
    \item ELECTR represents the electricity consumption monitoring of 370 customers over a span of three years. 
\end{enumerate}

The Table \ref{tab:short_term_full} demonstrates the performance of RWKV-TS in short-term time series forecasting. Although it does not achieve the best results overall, it is very close to several state-of-the-art (SOTA) models, with a small margin of difference.

\input{tables/short_term_full}

\subsection{Imputation Results}
\label{appendix:Imputation}

\paragraph{Datasets for Imputation} We use 6 datasets from long-term forecasting: Weather, ECL, ETTm1, ETTm2, ETTh1, and ETTh2. 

The table \ref{tab:imputation_full} provides a detailed demonstration of the performance of RWKV-TS in time series imputation task. 

\input{tables/imputation_full}

\subsection{Classification Results}
\label{appendix:Classification}

\paragraph{Datasets for Classification} Time series classification plays a crucial role in identification and medical diagnostics. To evaluate the model's ability to learn high-level representations, we undertake sequence-level classification. We select 10 diverse multivariate datasets from the UEA Time Series Classification Archive\cite{UEA}, encompassing gesture, action, audio recognition, and medical diagnosis via heartbeat monitoring, among other practical tasks. Following the preprocessing steps outlined by Zerveas(2020), we prepare the datasets, which vary in sequence length across different subsets. 

The table \ref{tab:classification_full} provides a full results of the performance of RWKV-TS in time series classification task. 

\input{tables/classification_full}

\subsection{Anomaly Detection Results}
\label{appendix:anomaly}

\paragraph{Datasets for Anomaly Detection} We benchmark our models against five well-established anomaly detection datasets:  SMD\cite{SMD}, MSL\cite{MSL_SMAP}, SMAP\cite{MSL_SMAP}, SWaT\cite{SWaT}, and PSM\cite{PSM}, covering applications in service monitoring, space and earth exploration, and water treatment.
Adhering to the preprocessing techniques outlined in \cite{xu2021anomaly}, we divide the dataset into consecutive, non-overlapping segments using a sliding window approach. In prior research, reconstruction has been a standard task for unsupervised point-wise representation learning, with reconstruction error serving as a natural anomaly detection metric. To ensure a fair comparison, we modify only the base models for reconstruction and utilize the classical reconstruction error as the common anomaly criterion across all experiments.

 The table \ref{tab:anomaly_full} offers a comprehensive overview of RWKV-TS's performance in the Time Series Anomaly Detection task.

\input{tables/anomaly_full}

\subsection{Efficiency Results}
\label{appendix:efficiency}

In Table \ref{tab:efficiency}, we have measured the memory usage and inference time for some models. We found that compared to the standard Transformer and its variants, RWKV-TS has less memory consumption and inference time. Although it has a larger memory footprint than purely linear models like DLinear, RWKV-TS delivers better performance. This suggests that RWKV-TS might offer a more favorable trade-off between effectiveness and efficiency. It would be particularly advantageous when deployed on low-resource end-devices. Of course, the comparison here is not entirely fair; we have simply followed the parameters used in previous work for our experiments, without fully aligning the parameter counts. 

\input{tables/efficiency}

%% file: tables/Long-term_full.tex
\begin{table*}[ht]
\begin{center}
\begin{small}
\scalebox{0.70}{
    \setlength\tabcolsep{3pt}
    \begin{tabular}{c|c|cc|cc|cc|cc|cc|cc|cc|cc|cc|cc|cc}
    \toprule
    
    \multicolumn{2}{c|}{Methods}&\multicolumn{2}{c|}{RWKV-TS}&\multicolumn{2}{c|}{DLinear}&\multicolumn{2}{c|}{PatchTST}&\multicolumn{2}{c|}{TimesNet}&\multicolumn{2}{c|}{FEDformer}&\multicolumn{2}{c|}{Autoformer}&\multicolumn{2}{c|}{Stationary}&\multicolumn{2}{c|}{ETSformer}&\multicolumn{2}{c|}{LightTS}&\multicolumn{2}{c|}{Informer}&\multicolumn{2}{c}{Reformer} \\
    
    \midrule
    
    \multicolumn{2}{c|}{Metric} & MSE & MAE& MSE & MAE& MSE  & MAE& MSE  & MAE& MSE  & MAE& MSE  & MAE& MSE  & MAE& MSE  & MAE& MSE  & MAE& MSE  & MAE& MSE  & MAE\\
    \midrule
    
    \multirow{5}{*}{\rotatebox{90}{$Weather$}}
    & 96  & 0.148 & 0.198 & 0.176 & 0.237 & 0.149 & 0.198 &0.172&0.220&0.217&0.296&0.266&0.336&0.173&0.223&0.197&0.281&0.182&0.242&0.300&0.384&0.689&0.596\\
    & 192 & 0.196 & 0.243 & 0.220 & 0.282 & 0.194 & 0.241 &0.219&0.261&0.276&0.336&0.307&0.367&0.245&0.285&0.237&0.312&0.227&0.287&0.598&0.544&0.752&0.638\\
    & 336 & 0.253 & 0.286 & 0.265 & 0.319 & 0.245 & 0.282&0.280&0.306&0.339&0.380&0.359&0.395&0.321&0.338&0.298&0.353&0.282&0.334&0.578&0.523&0.639&0.596 \\
    & 720 & 0.329 & 0.340 & 0.333 & 0.362 & 0.314 & 0.334&0.365&0.359&0.403&0.428&0.419&0.428&0.414&0.410&0.352&0.288&0.352&0.386&1.059&0.741&1.130&0.792\\
    & Avg &\color{red}\textbf{0.231}&\color{red}\textbf{0.266}&0.248&0.300&\textbf{0.225}&\textbf{0.264}&0.259&0.287&0.309&0.360&0.338&0.382&0.288&0.314&0.271&0.334&0.261&0.312&0.634&0.548&0.803&0.656\\
    \midrule
    
    \multirow{5}{*}{\rotatebox{90}{$ETTh1$}}
    & 96  & 0.384 & 0.414 & 0.375 & 0.399 & 0.370 & 0.399 &0.384&0.402&0.376&0.419&0.449&0.459&0.513&0.491&0.494&0.479&0.424&0.432&0.865&0.713&0.837&0.728\\
    & 192 & 0.415 & 0.433 & 0.405 & 0.416 & 0.413 & 0.421&0.436&0.429&0.420&0.448&0.500&0.482&0.534&0.504&0.538&0.504&0.475&0.462&1.008&0.792&0.923&0.766\\
    & 336 & 0.444 & 0.452 & 0.439 & 0.443 & 0.422 & 0.436 &0.491&0.469&0.459&0.465&0.521&0.496&0.588&0.535&0.574&0.521&0.518&0.488&1.107&0.809&1.097&0.835\\
    & 720 & 0.488 & 0.481 & 0.472 & 0.490 & 0.447 & 0.466 &0.521&0.500&0.506&0.507&0.514&0.512&0.643&0.616&0.562&0.535&0.547&0.533&1.181&0.865&1.257&0.889\\
    & Avg &0.433&0.445&\color{red}\textbf{0.422}&\color{red}\textbf{0.437}&\textbf{0.413}&\textbf{0.430}&0.458&0.450&0.440&0.460&0.496&0.487&0.570&0.537&0.542&0.510&0.491&0.479&1.040&0.795&1.029&0.805\\
    \midrule
    
    \multirow{5}{*}{\rotatebox{90}{$ETTh2$}}
    & 96  & 0.311 & 0.364 & 0.289 & 0.353 & 0.274 & 0.336&0.340&0.374&0.358&0.397&0.346&0.388&0.476&0.458&0.340&0.391&0.397&0.437&3.755&1.525&2.626&1.317 \\
    & 192 & 0.376 & 0.410 & 0.383 & 0.418 & 0.339 &  0.379 &0.402&0.414&0.429&0.439&0.456&0.452&0.512&0.493&0.430&0.439&0.520&0.504&5.602&1.931&11.12&2.979\\
    & 336 & 0.390 & 0.420 & 0.448 & 0.465 & 0.329 & 0.380&0.452&0.452&0.496&0.487&0.482&0.486&0.552&0.551&0.485&0.479&0.626&0.559&4.721&1.835&9.323&2.769 \\
    & 720 & 0.421 & 0.454 & 0.605 & 0.551 & 0.379 & 0.422&0.462&0.468&0.463&0.474&0.515&0.511&0.562&0.560&0.500&0.497&0.863&0.672&3.647&1.625&3.874&1.697\\
    & Avg &\color{red}\textbf{0.375}&\color{red}\textbf{0.412}&0.431&0.446&\textbf{0.330}&\textbf{0.379}&0.414&0.427&0.437&0.449&0.450&0.459&0.526&0.516&0.439&0.452&0.602&0.543&4.431&1.729&6.736&2.191\\
    \midrule
    
    \multirow{5}{*}{\rotatebox{90}{$ETTm1$}}
    & 96  & 0.302 & 0.356 & 0.299 & 0.343 & 0.290 & 0.342 &0.338&0.375&0.379&0.419&0.505&0.475&0.386&0.398&0.375&0.398&0.374&0.400&0.672&0.571&0.538&0.528 \\
    & 192 & 0.348 & 0.386 & 0.335 & 0.365 & 0.332 & 0.369&0.374&0.387&0.426&0.441&0.553&0.496&0.459&0.444&0.408&0.410&0.400&0.407&0.795&0.669&0.658&0.592\\
    & 336 & 0.391 & 0.412 & 0.369 & 0.386 & 0.366 & 0.392&0.410&0.411&0.445&0.459&0.621&0.537&0.495&0.464&0.435&0.428&0.438&0.438&1.212&0.871&0.898&0.721\\
    & 720 & 0.464 & 0.452 & 0.425 & 0.421 & 0.416 & 0.420&0.478&0.450&0.543&0.490&0.671&0.561&0.585&0.516&0.499&0.462&0.527&0.502&1.166&0.823&1.102&0.841\\
    & Avg&0.376&0.401&\color{red}\textbf{0.357}&\textbf{0.378}&\textbf{0.351}&\color{red}\textbf{0.380}&0.400&0.406&0.448&0.452&0.588&0.517&0.481&0.456&0.429&0.425&0.435&0.437&0.961&0.734&0.799&0.671 \\
    \midrule
    
    \multirow{5}{*}{\rotatebox{90}{$ETTm2$}}
    & 96  & 0.178 & 0.270 & 0.167 & 0.269 & 0.165 & 0.255&0.187&0.267&0.203&0.287&0.255&0.339&0.192&0.274&0.189&0.280&0.209&0.308&0.365&0.453&0.658&0.619\\
    & 192 & 0.251 & 0.314 & 0.224 & 0.303 & 0.220 & 0.292&0.249&0.309&0.269&0.328&0.281&0.340&0.280&0.339&0.253&0.319&0.311&0.382&0.533&0.563&1.078&0.827\\
    & 336 & 0.310 & 0.353 & 0.281 & 0.342 & 0.274 & 0.329 &0.321&0.351&0.325&0.366&0.339&0.372&0.334&0.361&0.314&0.357&0.442&0.466&1.363&0.887&1.549&0.972\\
    & 720 & 0.407 & 0.414 & 0.397 & 0.421 & 0.362 & 0.385&0.408&0.403&0.421&0.415&0.433&0.432&0.417&0.413&0.414&0.413&0.675&0.587&3.379&1.338&2.631&1.242 \\
    & Avg&0.287&0.338&\color{red}\textbf{0.267}&\color{red}\textbf{0.333}&\textbf{0.255}&\textbf{0.315}&0.291&0.333&0.305&0.349&0.327&0.371&0.306&0.347&0.293&0.342&0.409&0.436&1.410&0.810&1.479&0.915 \\
    \midrule
    
    \multirow{5}{*}{\rotatebox{90}{$ILI$}}
    & 24 &2.036 & 0.916 & 2.215 & 1.081 & 1.319 &0.754&2.317&0.934&3.228&1.260&3.483&1.287&2.294&0.945&2.527&1.020&8.313&2.144&5.764&1.677&4.400&1.382\\
    & 36 & 1.916 & 0.920 & 1.963 & 0.963 & 1.430 & 0.834&1.972&0.920&2.679&1.080&3.103&1.148&1.825&0.848&2.615&1.007&6.631&1.902&4.755&1.467&4.783&1.448 \\
    & 48 & 1.896 & 0.937 & 2.130 & 1.024 & 1.553 & 0.815 &2.238&0.940&2.622&1.078&2.669&1.085&2.010&0.900&2.359&0.972&7.299&1.982&4.763&1.469&4.832&1.465 \\
    & 60 & 1.790 & 0.927 & 2.368 & 1.096 & 1.470 & 0.788&2.027&0.928&2.857&1.157&2.770&1.125&2.178&0.963&2.487&1.016&7.283&1.985&5.264&1.564&4.882&1.483 \\
    & Avg &\color{red}\textbf{1.910}&\color{red}\textbf{0.925}&2.169&1.041&\textbf{1.443}&\textbf{0.797}&2.139&0.931&2.847&1.144&3.006&1.161&2.077&0.914&2.497&1.004&7.382&2.003&5.137&1.544&4.724&1.445\\
    \midrule
    
    \multirow{5}{*}{\rotatebox{90}{$ECL$}}
    & 96  & 0.129 & 0.224 &0.140 & 0.237 & 0.129 & 0.222&0.168&0.272&0.193&0.308&0.201&0.317&0.169&0.273&0.187&0.304&0.207&0.307&0.274&0.368&0.312&0.402 \\
    & 192 & 0.147 & 0.241 &0.153 & 0.249 & 0.157 &0.240&0.184&0.289&0.201&0.315&0.222&0.334&0.182&0.286&0.199&0.315&0.213&0.316&0.296&0.386&0.348&0.433\\
    & 336 & 0.163 & 0.258 &0.169 & 0.267 & 0.163 & 0.259&0.198&0.300&0.214&0.329&0.231&0.338&0.200&0.304&0.212&0.329&0.230&0.333&0.300&0.394&0.350&0.433\\
    & 720 & 0.199 & 0.291 &0.203 & 0.301 & 0.197 & 0.290&0.220&0.320&0.246&0.355&0.254&0.361&0.222&0.321&0.233&0.345&0.265&0.360&0.373&0.439&0.340&0.420\\
    & Avg &\textbf{0.159}&\textbf{0.253}&0.166&0.263&\color{red}\textbf{0.161}&\color{red}\textbf{0.253}&0.192&0.295&0.214&0.327&0.227&0.338&0.193&0.296&0.208&0.323&0.229&0.329&0.311&0.397&0.338&0.422\\
    \midrule
    
    \multirow{5}{*}{\rotatebox{90}{$Traffic$}}
    & 96  & 0.364 & 0.260 & 0.410 & 0.282 & 0.360 & 0.249&0.593&0.321&0.587&0.366&0.613&0.388&0.612&0.338&0.607&0.392&0.615&0.391&0.719&0.391&0.732&0.423\\
    & 192 & 0.385 & 0.269 &0.423 & 0.287 &0.379 & 0.256&0.617&0.336&0.604&0.373&0.616&0.382&0.613&0.340&0.621&0.399&0.601&0.382&0.696&0.379&0.733&0.420\\
    & 336 & 0.402 & 0.278 &0.436 & 0.296 & 0.392 & 0.264&0.629&0.336&0.621&0.383&0.622&0.337&0.618&0.328&0.622&0.396&0.613&0.386&0.777&0.420&0.742&0.420 \\
    & 720 & 0.442 & 0.298 &0.466 & 0.315 & 0.432 & 0.286&0.640&0.350&0.626&0.382&0.660&0.408&0.653&0.355&0.632&0.396&0.658&0.407&0.864&0.472&0.755&0.423\\
    & Avg & \color{red}\textbf{0.398}& \color{red}\textbf{0.276}&0.433&0.295&\textbf{0.390}&\textbf{0.263}&0.620&0.336&0.610&0.376&0.628&0.379&0.624&0.340&0.621&0.396&0.622&0.392&0.764&0.416&0.741&0.422\\
    \midrule
    \multicolumn{2}{c|}{Average}&\color{red}\textbf{0.521}&\color{red}\textbf{0.414}&0.562&0.436&\textbf{0.446}&\textbf{0.386}&0.596&0.433&0.701&0.489&0.757&0.511&0.633&0.465&0.662&0.473&1.303&0.616&1.836&0.871&2.081&0.954\\
    
    \bottomrule
    \end{tabular}
}
\end{small}
\end{center}
\caption{Full results on full data. We use prediction length $O \in \{96, 192, 336, 720\}$ for ILI and $O \in \{24, 36, 48, 60\}$ for others. A lower MSE indicates better performance. \textbf{Black}: best, {\textcolor{red}{Red}}: second best.}
\label{tab:100_percent_ett_full}
\end{table*}

%% file: tables/short_term_full.tex
\begin{table*}[ht]
\renewcommand\arraystretch{1.5}
\begin{center}
\begin{small}
\scalebox{0.7}{
\setlength\tabcolsep{3pt}
\begin{tabular}{cc|ccccccccccccc}
\toprule

\multicolumn{2}{c|}{Methods}&RWKV-TS&TimesNet&PatchTST&N-HiTS&N-BEATS& ETSformer& LightTS& DLinear &FEDformer &Stationary &Autoformer  &Informer&Reformer \\

% \multirow{2}{*}{Methods} 
% &\multicolumn{2}{c|}{GPT2(6)} & \multicolumn{2}{c|}{TimesNet}&\multicolumn{2}{c|}{ETSformer}&\multicolumn{2}{c|}{ETSformer}&\multicolumn{2}{c|}{LightTS}&\multicolumn{2}{c|}{DLinear}&\multicolumn{2}{c|}{FEDformer}&\multicolumn{2}{c|}{Stationary}&\multicolumn{2}{c|}{Autoformer}&\multicolumn{2}{c}{Informer}&\multicolumn{2}{c}{Reformer} \\

\midrule
\multirow{3}{*}{\rotatebox{90}{$Yearly$}}
&SMAPE&13.74&{\bf13.387}&13.477&13.418&13.436&18.009&14.247&16.965&13.728&13.717&13.974&14.727&16.169\\
&MASE&3.145&{\bf2.996}&3.019&3.045&3.043&4.487&3.109&4.283&3.048&3.078&3.134&3.418&3.800\\
&OWA&0.816&{\bf0.786}&0.792&0.793&0.794&1.115&0.827&1.058&0.803&0.807&0.822&0.881&0.973\\
\bottomrule

\multirow{3}{*}{\rotatebox{90}{$Quarterly$}}
&SMAPE&10.257&{\bf10.100}&10.38&10.202&10.124&13.376&11.364&12.145&10.792&10.958&11.338&11.360&13.313\\
&MASE&1.197&{\bf1.182}&1.233&1.194&1.169&1.906&1.328&1.520&1.283&1.325&1.365&1.401&1.775\\
&OWA&0.902&{\bf0.890}&0.921&0.899&0.886&1.302&1.000&1.106&0.958&0.981&1.012&1.027&1.252\\
\bottomrule

\multirow{3}{*}{\rotatebox{90}{$Monthly$}}
&SMAPE&12.793&{\bf12.670}&12.959&12.791&12.677&14.588&14.014&13.514&14.260&13.917&13.958&14.062&20.128\\
&MASE&0.941&{\bf0.933}&0.970&0.969&0.937&1.368&1.053&1.037&1.102&1.097&1.103&1.141&2.614\\
&OWA&0.886&{\bf0.878}&0.905&0.899&0.880&1.149&0.981&0.956&1.012&0.998&1.002&1.024&1.927\\
\bottomrule

\multirow{3}{*}{\rotatebox{90}{$Others$}}
&SMAPE&5.169&{\bf4.891}&4.952&5.061&4.925&7.267&15.880&6.709&4.954&6.302&5.485&24.460&32.491\\
&MASE&3.379&3.302&3.347&{\bf3.216}&3.391&5.240&11.434&4.953&3.264&4.064&3.865&20.960&33.355\\
&OWA&1.077&{\bf1.035}&1.049&1.040&1.053&1.591&3.474&1.487&1.036&1.304&1.187&5.879&8.679\\
\bottomrule

\multirow{3}{*}{\rotatebox{90}{$Average$}}
&SMAPE &12.021 &{\bf11.829}&12.059& 11.927& 11.851& 14.718& 13.525& 13.639 &12.840 &12.780 &12.909 &14.086 &18.200 \\
&MASE & 1.631 & {\bf1.585}&1.623 & 1.613 & 1.599 &2.408 &2.111 &2.095 &1.701 &1.756 &1.771  &2.718 &4.223\\
&OWA &0.870 & {\bf0.851}&0.869 &0.861 &0.855 &1.172 &1.051 &1.051 &0.918 &0.930 &0.939 & 1.230 & 1.775\\

\bottomrule

\end{tabular}
}
\end{small}
\end{center}
\caption{Full results of short-term forecasting.}
\label{tab:short_term_full}
\end{table*}

%% file: tables/imputation_full.tex
\begin{table}[ht]
\begin{center}
\begin{small}
\scalebox{0.65}{
    \setlength\tabcolsep{3pt}
    \begin{tabular}{cc|cc|cc|cc|cc|cc|cc|cc|cc|cc|cc|cc}
    \toprule
    
    \multicolumn{2}{c|}{Methods} 
    &\multicolumn{2}{c|}{RWKV-TS}&\multicolumn{2}{c|}{TimesNet}& \multicolumn{2}{c|}{PatchTST}&\multicolumn{2}{c|}{ETSformer}&\multicolumn{2}{c|}{LightTS}&\multicolumn{2}{c|}{DLinear}&\multicolumn{2}{c|}{FEDformer}&\multicolumn{2}{c|}{Stationary}&\multicolumn{2}{c|}{Autoformer}&\multicolumn{2}{c}{Informer}&\multicolumn{2}{c}{Reformer} \\
    Mask&Ratio&MSE&MAE&MSE&MAE&MSE&MAE&MSE&MAE&MSE&MAE&MSE&MAE&MSE&MAE&MSE&MAE&MSE&MAE&MSE&MAE&MSE&MAE \\
    
    \midrule
    \multirow{5}{*}{\rotatebox{90}{$ETTm1$}}
    & 12.5\% &0.102&0.183&0.023&0.101&0.041&0.130&0.096&0.229&0.093&0.206&0.080&0.193&0.052&0.166&0.032&0.119&0.046&0.144&0.063&0.180&0.042&0.146 \\
    & 25\% &0.131&0.207&0.023&0.101&0.044&0.135&0.096&0.229&0.093&0.206&0.080&0.193&0.052&0.166&0.032&0.119&0.046&0.144&0.063&0.180&0.042&0.146 \\
    & 37.5\% &0.165&0.231&0.029&0.111&0.049&0.143&0.133&0.271&0.113&0.231&0.103&0.219&0.069&0.191&0.039&0.131&0.057&0.161&0.079&0.200&0.063&0.182 \\
    & 50\% &0.199&0.255&0.036&0.124&0.055&0.151&0.186&0.323&0.134&0.255&0.132&0.248&0.089&0.218&0.047&0.145&0.067&0.174&0.093&0.218&0.082&0.208 \\
    & Avg &0.149&0.219&0.027&0.107&0.047&0.140&0.120&0.253&0.104&0.218&0.093&0.206&0.062&0.177&0.036&0.126&0.051&0.150&0.071&0.188&0.055&0.166 \\
    \midrule
    
    \multirow{5}{*}{\rotatebox{90}{$ETTm2$}}
    & 12.5\% &0.025&0.100&0.018&0.080&0.026&0.094&0.108&0.239&0.034&0.127&0.062&0.166&0.056&0.159&0.021&0.088&0.023&0.092&0.133&0.270&0.108&0.228 \\
    & 25\% &0.028&0.108&0.020&0.085&0.028&0.099&0.164&0.294&0.042&0.143&0.085&0.196&0.080&0.195&0.024&0.096&0.026&0.101&0.135&0.272&0.136&0.262 \\
    & 37.5\% &0.031&0.114&0.023&0.091&0.030&0.104&0.237&0.356&0.051&0.159&0.106&0.222&0.110&0.231&0.027&0.103&0.030&0.108&0.155&0.293&0.175&0.300 \\
    & 50\% &0.036&0.123&0.026&0.098&0.034&0.110&0.323&0.421&0.059&0.174&0.131&0.247&0.156&0.276&0.030&0.108&0.035&0.119&0.200&0.333&0.211&0.329 \\
    & Avg &0.030&0.111&0.022&0.088&0.029&0.102&0.208&0.327&0.046&0.151&0.096&0.208&0.101&0.215&0.026&0.099&0.029&0.105&0.156&0.292&0.157&0.280 \\
    \midrule
    
    \multirow{5}{*}{\rotatebox{90}{$ETTh1$}}
    & 12.5\% &0.204&0.272&0.057&0.159&0.093&0.201&0.126&0.263&0.240&0.345&0.151&0.267&0.070&0.190&0.060&0.165&0.074&0.182&0.114&0.234&0.074&0.194 \\
    & 25\% &0.239&0.295&0.069&0.178&0.107&0.217&0.169&0.304&0.265&0.364&0.180&0.292&0.106&0.236&0.080&0.189&0.090&0.203&0.140&0.262&0.102&0.227 \\
    & 37.5\% &0.293&0.324&0.084&0.196&0.120&0.230&0.220&0.347&0.296&0.382&0.215&0.318&0.124&0.258&0.102&0.212&0.109&0.222&0.174&0.293&0.135&0.261 \\
    & 50\%&0.335&0.350&{\bf0.102}&{\bf0.215}&0.141&0.248&0.293&0.402&0.334&0.404&0.257&0.347&0.165&0.299&0.133&0.240&0.137&0.248&0.215&0.325&0.179&0.298 \\
    & Avg &0.268&0.310&0.078&0.187&0.115&0.224&0.202&0.329&0.284&0.373&0.201&0.306&0.117&0.246&0.094&0.201&0.103&0.214&0.161&0.279&0.122&0.245 \\
    \midrule
    
    \multirow{5}{*}{\rotatebox{90}{$ETTh2$}}
    & 12.5\% &0.049&0.148&0.040&0.130&0.057&0.152&0.187&0.319&0.101&0.231&0.100&0.216&0.095&0.212&0.042&0.133&0.044&0.138&0.305&0.431&0.163&0.289 \\
    & 25\% &0.054&0.157&0.046&0.141&0.061&0.158&0.279&0.390&0.115&0.246&0.127&0.247&0.137&0.258&0.049&0.147&0.050&0.149&0.322&0.444&0.206&0.331 \\
    & 37.5\% &0.063&0.170&0.052&0.151&0.067&0.166&0.400&0.465&0.126&0.257&0.158&0.276&0.187&0.304&0.056&0.158&0.060&0.163&0.353&0.462&0.252&0.370 \\
    & 50\% &0.074&0.184&0.060&0.162&0.073&0.174&0.602&0.572&0.136&0.268&0.183&0.299&0.232&0.341&0.065&0.170&0.068&0.173&0.369&0.472&0.316&0.419 \\
    & Avg &0.060&0.164&0.049&0.146&0.065&0.163&0.367&0.436&0.119&0.250&0.142&0.259&0.163&0.279&0.053&0.152&0.055&0.156&0.337&0.452&0.234&0.352 \\
    \midrule
    
    \multirow{5}{*}{\rotatebox{90}{$ECL$}}
    & 12.5\% &0.175&0.287&0.085&0.202&0.055&0.160&0.196&0.321&0.102&0.229&0.092&0.214&0.107&0.237&0.093&0.210&0.089&0.210&0.218&0.326&0.190&0.308 \\
    & 25\% &0.178&0.291&0.089&0.206&0.065&0.175&0.207&0.332&0.121&0.252&0.118&0.247&0.120&0.251&0.097&0.214&0.096&0.220&0.219&0.326&0.197&0.312 \\
    & 37.5\%  &0.182&0.295&0.094&0.213&0.076&0.189&0.219&0.344&0.141&0.273&0.144&0.276&0.136&0.266&0.102&0.220&0.104&0.229&0.222&0.328&0.203&0.315\\
    & 50\% &0.185&0.299&{\bf0.100}&0.221&0.091&0.208&0.235&0.357&0.160&0.293&0.175&0.305&0.158&0.284&0.108&0.228&0.113&0.239&0.228&0.331&0.210&0.319 \\
    & Avg &0.180&0.293&0.092&0.210&0.072&0.183&0.214&0.339&0.131&0.262&0.132&0.260&0.130&0.259&0.100&0.218&0.101&0.225&0.222&0.328&0.200&0.313 \\
    \midrule
    
    \multirow{5}{*}{\rotatebox{90}{$Weather$}}
    & 12.5\%& 0.031&0.058&{\bf0.025}&{\bf0.045}&0.029&0.049&0.057&0.141&0.047&0.101&0.039&0.084&0.041&0.107&0.027&0.051&0.026&0.047&0.037&0.093&0.031&0.076 \\
    & 25\% &0.039&0.061&0.029&0.052&0.031&0.053&0.065&0.155&0.052&0.111&0.048&0.103&0.064&0.163&0.029&0.056&0.030&0.054&0.042&0.100&0.035&0.082 \\
    & 37.5\% &0.040&0.060&{\bf0.031}&{\bf0.057}&0.035&0.058&0.081&0.180&0.058&0.121&0.057&0.117&0.107&0.229&0.033&0.062&0.032&0.060&0.049&0.111&0.040&0.091 \\
    & 50\% &0.044&0.073&{\bf0.034}&{\bf0.062}&0.038&0.063&0.102&0.207&0.065&0.133&0.066&0.134&0.183&0.312&0.037&0.068&0.037&0.067&0.053&0.114&0.046&0.099 \\
    & Avg &0.038&0.066&{\bf0.030}&{\bf0.054}&0.060&0.144&0.076&0.171&0.055&0.117&0.052&0.110&0.099&0.203&0.032&0.059&0.031&0.057&0.045&0.104&0.038&0.087 \\
    \bottomrule
    
    \end{tabular}
}
\end{small}
\end{center}
\caption{Full results for the imputation task.}
\label{tab:imputation_full}
\end{table}

%% file: tables/classification_full.tex
\begin{table*}[htbp]
\begin{center}
\begin{small}
\scalebox{0.7}{
    \setlength\tabcolsep{3pt}
    \begin{tabular}{c|cccc|cc|ccccccccc|cc|cc}
    \hline
    
    \multirow{2}{*}{Methods} & \multicolumn{4}{c|}{RNN} &
    \multicolumn{2}{c|}{Classical methods} & \multicolumn{9}{c|}{Transformers} & \multicolumn{2}{c|}{MLP} & \multicolumn{2}{c}{CNN} \\
    &RWKV-TS&LSTM&LSTNet& LSSL&XGBoost&Rocket&Trans.& Re.& In.& Pyra.& Auto.& Station.& FED.& ETS.& Flow. &DLinear &LightTS. & TimesNet & TCN\\
    
    \hline
    EthanolConcentration&34.6&32.3&43.7 &45.2 &39.9 &31.1 &32.7 &31.9 &31.6 &30.8 &31.6 &32.7 &31.2 &28.1 &33.8 &32.6 &29.7 &35.7&28.9\\
    FaceDetection&67.3&57.7&63.3 &64.7 &65.7 &66.7&67.3 &68.6 &67.0 &65.7 &68.4 &68.0 &66.0 &66.3 &67.6 &68.0 &67.5 &68.6&52.8\\
    Handwriting&34.6&15.2&15.8 &58.8 &25.8 &24.6&32.0 &27.4 &32.8 &29.4 &36.7 &31.6 &28.0 &32.5 &33.8 &27.0 &26.1 &32.1&53.3 \\
    Heartbeat&77.1&72.2&73.2 &75.6 &77.1 &72.7&76.1 &77.1 &80.5 &75.6 &74.6 &73.7 &73.7 &71.2 &77.6 &75.1 &75.1 &78.0 &75.6 \\
    JapaneseVowels&98.4&79.7&86.5 &96.2 &98.1 &98.4 &98.7 &97.8 &98.9 &98.4 &96.2 &99.2 &98.4 &95.9 &98.9 &96.2 &96.2 &98.4 &98.9 \\
    PEMS-SF&83.8&39.9&98.3 &75.1 &86.7 &86.1 &82.1 &82.7 &81.5 &83.2 &82.7 &87.3 &80.9 &86.0 &83.8 &75.1 &88.4 &89.6&68.8 \\
    SelfRegulationSCP1&91.8&68.9&84.6 &90.8 &84.0 &90.8 &92.2 &90.4 &90.1 &88.1 &84.0 &89.4 &88.7 &89.6 &92.5 &87.3 &89.8 &91.8&84.6\\
    SelfRegulationSCP2&57.2&46.6&48.9 &53.3 &52.8 &52.2 &53.9 &56.7 &53.3 &53.3 &50.6 &57.2 &54.4 &55.0 &56.1 &50.5 &51.1 &57.2&55.6 \\
    SpokenArabicDigits&98.9&31.9&69.6 &71.2 &100.0 &100.0&98.4 &97.0 &100.0 &99.6 &100.0 &100.0 &100.0 &100.0 &98.8 &81.4 &100.0 &99.0&95.6 \\
    UWaveGestureLibrary&86.9&41.2&75.9 &94.4 &87.8 &85.9&85.6 &85.6 &85.6 &83.4 &85.9 &87.5 &85.3 &85.0 &86.6 &82.1 &80.3 &85.3&88.4 \\
    \hline
    Average &\color{red}\textbf{73.1}&48.6&66.0 &72.5 &71.8 &70.9 &71.9 &71.5 &72.1 
    &70.8 &71.1 &72.7 &70.7 &71.0 &73.0 &67.5 &70.4 &\textbf{73.6}&70.3 \\
    \hline
    
    \end{tabular}
}
\end{small}
\end{center}
\vskip -0.1in
\caption{Full results for the classification task. $\ast$. in the Transformers indicates the name of $\ast$former.}
\label{tab:classification_full}
\end{table*}

%% file: tables/anomaly_full.tex
\begin{table}[htbp]
\vskip 0.15in
\begin{center}
\begin{small}
\scalebox{0.65}{
    \begin{tabular}{c|ccc|ccc|ccc|ccc|ccc|c}
    \toprule
    
    Methods &
    \multicolumn{3}{c|}{SMD} & \multicolumn{3}{c|}{MSL} & \multicolumn{3}{c|}{SMAP}& \multicolumn{3}{c|}{SWaT} &\multicolumn{3}{c|}{PSM} & Avg F1 \\
    Metrics&P&R&F1&P&R&F1&P&R&F1&P&R&F1&P&R&F1&\%  \\
    
    \midrule
    RWKV-TS&87.45&81.43&84.33&78.11&77.74&77.92&89.04&55.94&68.71&88.20&94.85&91.40&98.32&95.92&97.10&83.89\\
    TimesNet &87.91&81.54&84.61&{\bf89.54}&75.36&81.84&90.14&56.40&69.39&90.75&95.40&93.02&98.51&{\bf96.20}&{\bf97.34}&85.24\\
    PatchTST&87.26&82.14&84.62&88.34&70.96&78.70&90.64&55.46&68.82&91.10&80.94&85.72&98.84&93.47&96.08&82.79\\
    ETSformer&87.44&79.23&83.13&85.13&84.93&85.03&92.25&55.75&69.50&90.02&80.36&84.91&99.31&85.28&91.76&82.87\\
    FEDformer&87.95&82.39&85.08&77.14&80.07&78.57&90.47&58.10&70.76&90.17&96.42&93.19&97.31&97.16&97.23&84.97\\
    LightTS&87.10&78.42&82.53&82.40&75.78&78.95&92.58&55.27&69.21&91.98&94.72&93.33&98.37&95.97&97.15&84.23 \\
    DLinear&83.62&71.52&77.10&84.34&85.42&84.88&92.32&55.41&69.26&80.91&95.30&87.52&98.28&89.26&93.55&82.46 \\
    Stationary&88.33&81.21&84.62&68.55&89.14&77.50&89.37&59.02&71.09&68.03&96.75&79.88&97.82&96.76&97.29&82.08 \\
    Autoformer&88.06&82.35&85.11&77.27&80.92&79.05&90.40&58.62&71.12&89.85&95.81&92.74&99.08&88.15&93.29&84.26 \\
    Pyraformer&85.61&80.61&83.04&83.81&85.93&84.86&92.54&57.71&71.09&87.92&96.00&91.78&71.67&96.02&82.08&82.57 \\
    Anomaly Transformer&88.91&82.23&85.49&79.61&87.37&83.31&91.85&58.11&71.18&72.51&97.32&83.10&68.35&94.72&79.40&80.50 \\
    Informer&86.60&77.23&81.65&81.77&86.48&84.06&90.11&57.13&69.92&70.29&96.75&81.43&64.27&96.33&77.10&78.83 \\
    Reformer&82.58&69.24&75.32&85.51&83.31&84.40&90.91&57.44&70.40&72.50&96.53&82.80&59.93&95.38&73.61&77.31 \\
    LogTransformer&83.46&70.13&76.21&73.05&87.37&79.57&89.15&57.59&69.97&68.67&97.32&80.52&63.06&98.00&76.74&76.60 \\
    Transformer&83.58&76.13&79.56&71.57&87.37&78.68&89.37&57.12&69.70&68.84&96.53&80.37&62.75&96.56&76.07&76.88 \\
    
    \bottomrule
    
    \end{tabular}
}
\end{small}
\end{center}
\caption{Full results for the anomaly detection.}
\label{tab:anomaly_full}
\end{table}

%% file: tables/efficiency.tex
\begin{table}[htbp]
\vspace{-0.1cm}
        \centering
        \scalebox{0.9}
        % \scriptsize
        {
        	{%
                \begin{tabular}{c|cccc} \hline
                Method   & Parameter & Time & Memory\\\hline
                RWKV-TS  & 1.0M    & 3.0ms     &1560MiB      \\\hline
                DLinear  & 139.7K    & 0.4ms     &687MiB        \\\hline
                Transformer& 13.61M     & 26.8ms     &6091MiB \\
                Informer    & 14.39M     & 49.3ms     &3869MiB        \\
                Autoformer  & 14.91M     & 164.1ms     &7607MiB        \\
                FEDformer   & 20.68M     & 40.5ms     &4143MiB \\\hline               
                \end{tabular}%
        	}
            }
            \vspace{-0.2cm}
        \caption{Comparison of practical efficiency of different models under L=96 and T=720 on the Electricity. The inference time averages 5 runs.}
\label{tab:efficiency}
\vspace{-0.3cm}
\end{table}